\documentclass{article}

\usepackage{arxiv}
\usepackage[utf8]{inputenc}
\usepackage[T1]{fontenc}
\usepackage{amsmath,amssymb,amsfonts}
\usepackage{graphicx}
\usepackage{booktabs}
\usepackage{multirow}
\usepackage{makecell}
\usepackage[numbers,sort&compress]{natbib}
\usepackage{microtype}
\usepackage{authblk}
\usepackage[hidelinks]{hyperref}
\usepackage{url}

\title{NSF-HRPT: Neural Semantic Field meets Hierarchical Risk Perception Tree for Safety-Critical Scenario Assessment}

\author[1]{Yu Zhao}
\author[1]{Jiangyu Pan}
\author[1]{Tao Hu}
\author[1]{Ming Yin}
\author[1]{Fan Yang}
\author[2]{Jiangfan Liu}
\author[1]{Xiubo Liang\thanks{Corresponding author: Xiubo Liang, e-mail: \texttt{xiubo@zju.edu.cn}}}
\affil[1]{Zhejiang University, Ningbo, China}
\affil[2]{Beihang University, Beijing, China}

\date{}
\renewcommand{\shorttitle}{NSF-HRPT}
\renewcommand{\headeright}{Preprint}
\renewcommand{\undertitle}{Preprint}
\hypersetup{
  pdftitle={NSF-HRPT: Neural Semantic Field meets Hierarchical Risk Perception Tree for Safety-Critical Scenario Assessment},
  pdfauthor={Yu Zhao, Jiangyu Pan, Tao Hu, Ming Yin, Fan Yang, Jiangfan Liu, Xiubo Liang},
  pdfsubject={cs.AI},
  pdfkeywords={Safety-Critical Scenarios, Neural Semantic Fields, Sim-to-Real Transfer, Autonomous Driving, Embodied Intelligence}
}
\begin{document}

\maketitle

\begin{abstract}

The ability to accurately assess and anticipate risks in safety-critical scenarios is crucial for autonomous driving systems. While existing research has made progress in collision prediction, accurately quantifying risk levels from monocular vision inputs remains challenging due to the complex dynamics of multi-agent interactions and the inherent uncertainty in real-world environments. To address these challenges, we present \textbf{NSF-HRPT}, a novel framework that combines learning-based perception with structured reasoning for quantitative risk assessment. Our approach features a Neural Semantic Field (NSF) that learns to model scene semantics, trajectory predictions, and probabilistic Time-to-Collision (TTC) distributions from simulation data. During inference, the pre-trained NSF serves as a prior for our Hierarchical Risk Perception Tree (HRPT), which enables efficient parallel computation and spatial reasoning about multi-agent risks. Additionally, we introduce a Sim2Real enhancement strategy that improves real-world applicability without retraining by incorporating priors from foundation models. Extensive evaluations demonstrate that our framework achieves state-of-the-art performance on synthetic benchmarks and delivers competitive, near-state-of-the-art results on real-world datasets for both TTC estimation accuracy and risk localization precision. The proposed method provides an effective solution for real-time risk awareness from monocular camera inputs.
\end{abstract}


\section{INTRODUCTION}

The validation of autonomous driving systems is fundamentally benchmarked against their ability to navigate safety-critical scenarios---infrequent yet high-risk situations that define the operational boundaries of system robustness~\cite{ding2023survey, breitenstein2021corner}. This endeavor faces a significant data dichotomy. While large-scale datasets such as Waymo Open~\cite{sun2020scalability, Kan_2024_icra} and nuScenes~\cite{caesar2020nuscenes} offer extensive annotations for nominal driving conditions, they remain inherently sparse in critical incidents. Specialized accident datasets, including DAD~\cite{chan2016anticipating} and CCD~\cite{bao2020uncertainty}, provide valuable collision examples with Time-to-Collision (TTC) ground truth, yet their single-perspective, egocentric view lacks the comprehensive bird's-eye-view (BEV) trajectory information essential for precise multi-agent dynamic risk analysis.

To address these challenges, the research community has pursued two complementary directions: safety-critical scenario generation for offline validation~\cite{liu2024safety, xu2025diffscene, zhang2024chatscene}, and safety-critical scenario perception for real-time risk identification. Within the latter, previous works have primarily focused on predicting scene-level metrics such as mean-Time-to-Accident (mTTA), which estimates the probability of future collisions~\cite{karim2022dynamic, guan2025world}. However, for actionable decision-making in autonomous systems, a more fundamental and valuable capability is the precise estimation of Time-to-Collision (TTC) for individual entities, providing a continuous, physically-grounded measure of collision urgency. This capability is particularly crucial for embodied intelligent systems, such as autonomous vehicles, that must make reliable safety decisions based on limited visual observations.

This task of quantitative risk assessment from monocular vision presents several fundamental challenges: First, obtaining the comprehensive ground truth data necessary for training deep dynamics models, particularly precise BEV trajectories of all traffic participants, remains difficult in real-world settings. Second, performing efficient risk assessment for numerous agents in complex scenes requires computational frameworks that can scale effectively for real-time operation. Third, bridging the visual domain gap between simulation-trained models and real-world application demands robust adaptation strategies.

To address these challenges, we propose \textbf{NSF-HRPT}, a novel framework for structured risk reasoning from sequential monocular frames. Our approach leverages the capability of high-fidelity simulation platforms like CARLA~\cite{pmlr-v78-dosovitskiy17a} to generate diverse critical scenarios with comprehensive ground truth, including precise BEV trajectories and TTC values. The framework operates through two coordinated stages: an offline training phase that learns a powerful multi-task \textbf{Neural Semantic Field (NSF)} representing scene semantics, motion, and risk distributions; and an online inference phase where the pre-trained NSF drives a \textbf{Hierarchical Risk Perception Tree (HRPT)} for parallel, structured risk reasoning. Additionally, we incorporate a \textbf{Sim2Real Enhancement} strategy that bridges the visual domain gap without retraining by integrating geometric and semantic priors from foundation models.

The main contributions of our work are as follows:
\begin{enumerate}
    \item We propose a \textbf{Neural Semantic Field (NSF)} that learns a unified representation of scene semantics, motion, and risk from simulation data, with explicit probabilistic modeling of TTC uncertainty.
    \item We design a \textbf{Hierarchical Risk Perception Tree (HRPT)}, an efficient parallel inference algorithm that uses the pre-trained NSF for structured multi-agent risk reasoning in real time.
    \item We introduce a \textbf{Sim2Real Enhancement} strategy that leverages foundation models to bridge the visual domain gap without retraining, improving real-world applicability.
    \item We conduct extensive experiments in simulated and real-world safety-critical scenarios, showing our framework achieves state-of-the-art or near-state-of-the-art performance.
\end{enumerate}

\section{RELATED WORK}

\textbf{Safety-critical Scenario Generation} focuses on creating diverse test cases for offline evaluation, largely driven by generative models. High-fidelity simulators, such as CARLA \cite{pmlr-v78-dosovitskiy17a}, provide a critical execution environment for diverse generation algorithms, enabling evaluation in virtual worlds that approximate real-world physics and traffic rules. A prominent trend is the use of diffusion models to generate rare and realistic critical events \cite{xu2025diffscene}, create entire scenes with static road layouts and dynamic actors \cite{zhou2024diffroad}, or synthesize training data from semantic descriptions \cite{guan2025world}. Other strategies include autoregressive models for long-term traffic flows \cite{feng2022trafficgen} and reinforcement learning to iteratively edit scenarios for maximal risk while maintaining plausibility \cite{liu2024safety, yang2025authsim}. Additionally, LLM agents are increasingly used to interpret natural language into executable simulator instructions \cite{zhang2024chatscene} or augment standard scenarios with adversarial conditions \cite{li2025safety2drive}. Retrieval-augmented LLMs can further discover and memorize adversarial behaviors online \cite{mei2025seeking}, while knowledge-grounded LLMs reason about core threats amplified through multi-agent trajectory optimization \cite{liu2025adversarial}.

\textbf{Safety-Critical Scenarios Prediction} aims to identify hazards and anticipate accidents from sensor data, supported by datasets such as CCD \cite{bao2020uncertainty}, DAD \cite{chan2016anticipating}, A3D \cite{yao2019a3d-icra,yao2019a3d-iros}, and benchmarks for rare event perception \cite{bogdoll2024anovox, nekrasov2025spotting}. Methodologies increasingly emphasize handling open-world risks, including unsupervised approaches using generative world models to identify prediction anomalies \cite{bogdoll2024umad} and frameworks leveraging synthetic data to detect unseen objects with minimal real labels \cite{chen2025jisam}. While earlier work emphasized scene-level metrics using spatio-temporal attention or graph networks \cite{karim2022dynamic, thakur2024graph, liao2024crash}, recent efforts shift toward agent-specific risk understanding \cite{zhang2025latte}. This requires accurate egocentric trajectory prediction, motivating diffusion-based frameworks and visual memory models \cite{bae2024singulartrajectory, wang2024egonav, rasouli2024novel}. Concurrently, new methods target direct agent-level risk assessment, such as vision-language models for motion risk prediction \cite{hou2025drivemrp} and LLMs to identify agents involved in predicted collisions \cite{liao2024and}.

\section{METHOD}

We propose \textbf{NSF-HRPT}, a novel framework for structured risk reasoning from monocular video in safety-critical street scenes. It operates through two coordinated components: a trainable \textbf{Neural Semantic Field (NSF)} that learns a unified representation of scene semantics, motion, and risk, and a \textbf{Hierarchical Risk Perception Tree (HRPT)} that enables efficient parallel risk assessment at inference time using the pre-trained NSF. A \textbf{Sim2Real enhancement} strategy further bridges the sim-to-real visual domain gap without retraining.

\begin{figure}[htbp]
  \centering
  \includegraphics[width=\linewidth]{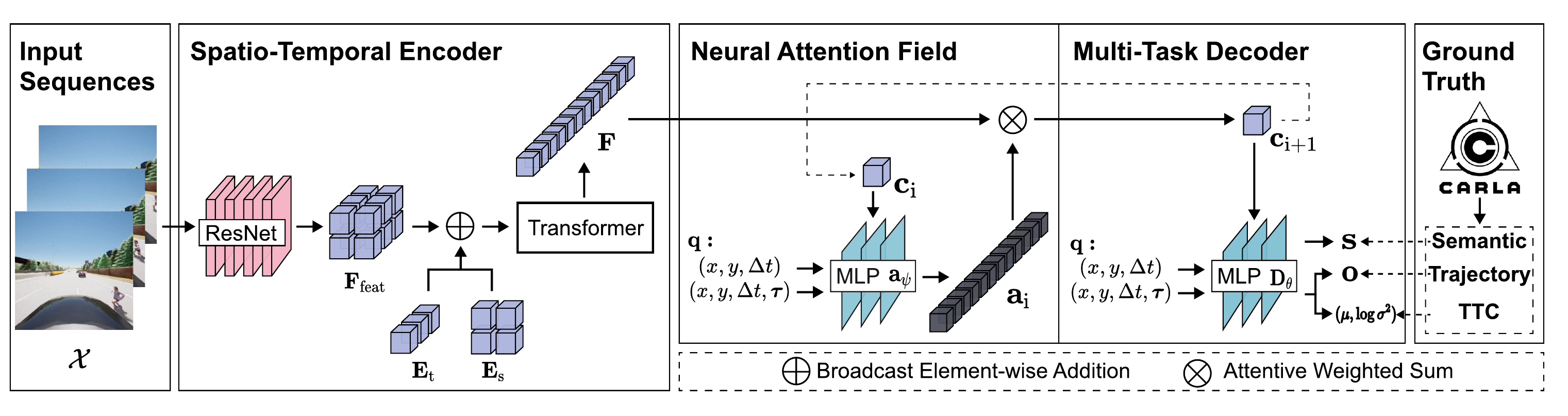}
  \caption{\textbf{Neural Semantic Field (NSF) Architecture.} The NSF model comprises three core modules: a Spatio-Temporal Encoder for feature extraction, a Neural Attention Field for iterative query-conditioned context refinement, and a Multi-task Decoder that outputs semantic occupancy, trajectory offsets, and probabilistic TTC distributions. The illustration uses \( T=3 \) input frames and \( P=4 \) spatial patches for demonstration.}
\label{fig:nsf_architecture}
\end{figure}

\subsection{Neural Semantic Field (NSF)}
Inspired by NEAT~\cite{Chitta2021ICCV}, the Neural Semantic Field (NSF) is a continuous representation that maps queries from monocular multi-frame images to semantic logits, trajectory, and TTC risk predictions. The Neural Semantic Field (NSF) comprises three core components: a Spatio-Temporal Encoder, a Neural Attention Field, and a Multi-task Decoder, as illustrated in Figure~\ref{fig:nsf_architecture}.

\textbf{Spatio-Temporal Encoder.}
The Spatio-Temporal Encoder \( \mathbf{E}_{\phi} \) extracts spatio-temporal features from an input sequence \( \mathcal{X} \in \mathbb{R}^{T \times H \times W \times 3} \) representing multi-frame images. A ResNet backbone~\cite{he2016deep} first extracts a feature tensor \( \mathbf{F}_{\text{feat}} \in \mathbb{R}^{T \times P \times C} \), where \( P=64 \) is the number of spatial patches and \( C \) is the feature dimensionality. Learnable spatial \( \mathbf{E}_s \in \mathbb{R}^{P \times C} \) and temporal \( \mathbf{E}_t \in \mathbb{R}^{T \times C} \) positional embeddings are broadcast and added element-wise .
\begin{equation}
\mathbf{F}_{\text{feat}}'[t, p, :] = \mathbf{F}_{\text{feat}}[t, p, :] + \mathbf{E}_s[p, :] + \mathbf{E}_t[t, :].
\end{equation}
The tensor \(\mathbf{F}_{\text{feat}}'\) is then processed by a transformer~\cite{ATTISAYN}, which globally integrates spatio-temporal features via self-attention, enriching each patch with contextual information across space and time. The resulting tensor is flattened into the final feature matrix \( \mathbf{F} \in \mathbb{R}^{(T \cdot P) \times C} \).

\textbf{Neural Attention Field.}
The Neural Attention Field \( \mathbf{a}_{\psi} \) implements iterative, query-dependent attention. It operates on the feature matrix \( \mathbf{F} \) and a query vector \( \mathbf{q} \), which can be either \( \mathbf{q}_1 = (x, y, \Delta t) \) for semantic occupancy prediction or \( \mathbf{q}_2 = (x, y, \Delta t, \boldsymbol{\tau}) \) for behavior prediction, where \( (x, y) \) is a BEV coordinate, \( \Delta t \) is a future time offset, and \( \boldsymbol{\tau} \in \mathbb{R}^{D_\tau} \) is a learnable entity-type embedding which represents distinct behavioral priors for different entity categories (e.g., vehicles, pedestrians). The module initializes a context vector \( \mathbf{c}_0 \) to the mean of \( \mathbf{F} \). For \( N \) steps, an MLP \( a_{\psi} \) computes an attention weight vector:
\begin{equation}
\mathbf{a}_i = a_{\psi}([\mathbf{q}, \mathbf{c}_i]),
\end{equation}
used to produce the next context vector:
\begin{equation}
\mathbf{c}_{i+1} = \text{softmax}(\mathbf{a}_i)^{\top} \mathbf{F}.
\end{equation}
The final output is the refined context \( \mathbf{c}_N \in \mathbb{R}^{C} \).

\textbf{Multi-task Decoder.}
The Multi-task Decoder \( \mathbf{D_{\theta}} \) is implemented as a MLP that processes the concatenated input \( [\mathbf{q}, \mathbf{c}_N] \in \mathbb{R}^{(D_q + C)} \), where \( D_q \) represents the dimensionality of the query vector and varies depending on the query type (3 for \(\mathbf{q}_1\), \(3 + D_{\tau}\) for \(\mathbf{q}_2\)). The decoder produces specialized outputs through two distinct mapping functions:

\textit{Semantic Occupancy Prediction:} 
\begin{equation}
    \mathbf{q}_1: (x, y, \Delta t) \mapsto \mathbf{s},
\end{equation}
predicts the semantic class distribution \( \mathbf{s} \in \mathbb{R}^{6} \) (vehicle, pedestrian, road, red light, green light, background) at the fixed coordinate \( (x, y) \) after time interval \( \Delta t \).

\textit{Entity Behavior Prediction:} 
\begin{equation}
    \mathbf{q}_2: (x, y, \Delta t, \boldsymbol{\tau}) \mapsto \mathbf{o},\ (\mu, \log \sigma^2),
\end{equation}
predicts the future state of the entity currently located at \( (x, y) \). It outputs both the trajectory offset vector \( \mathbf{o} \in \mathbb{R}^{2} \) and the TTC distribution parameters, comprising the mean \( \mu \in \mathbb{R} \) and log variance \( \log \sigma^2 \in \mathbb{R} \) which quantifies prediction uncertainty, after time interval \( \Delta t \).

The decoder employs Conditional Batch Normalization (CBN), with parameters dynamically modulated based on the input query features to specialize the processing for each task.

\textbf{Multi-task Loss.}
The model is trained end-to-end by minimizing the objective:
\begin{equation}
\mathcal{L}_{\text{total}} = \lambda_{\text{sem}} \mathcal{L}_{\text{CE}} + \lambda_{\text{traj}} \mathcal{L}_{\text{SmoothL1}} + \lambda_{\text{ttc}} \mathcal{L}_{\text{NLL}}.
\end{equation}
The constituent losses are defined as follows. The semantic loss is the cross-entropy:
\begin{equation}
\mathcal{L}_{\text{CE}} = -\sum_{k=1}^{K} s_k^{\text{gt}} \log \left( \frac{\exp(s_k)}{\sum_{j=1}^{K} \exp(s_j)} \right).
\end{equation}
The offset loss uses the Smooth L1 norm:
\begin{equation}
\mathcal{L}_{\text{SmoothL1}} = 
\begin{cases} 
0.5 (\mathbf{o} - \mathbf{o}^{\text{gt}})^2 / \beta, & \text{if } |\mathbf{o} - \mathbf{o}^{\text{gt}}| < \beta, \\ 
|\mathbf{o} - \mathbf{o}^{\text{gt}}| - 0.5 \beta, & \text{otherwise},
\end{cases}
\end{equation}
where \( \beta = 1.0 \) is a predefined threshold parameter. 

The TTC loss is the negative log-likelihood under a Gaussian distribution:
\begin{equation}
\mathcal{L}_{\text{NLL}} = \frac{1}{2} (TTC^{\text{gt}} - \mu)^2 \exp(-\log \sigma^2) + \frac{1}{2} \log \sigma^2.
\end{equation}
Hyperparameters \( \lambda_{\text{sem}}, \lambda_{\text{traj}}, \lambda_{\text{ttc}} \) balance the task weights.

The training process utilizes ground truth data from the CARLA simulation environment~\cite{pmlr-v78-dosovitskiy17a} to optimize the parameters $\Theta = \{\phi, \psi, \theta\}$ and learnable embeddings $\mathbf{E}_s$, $\mathbf{E}_t$, and $\boldsymbol{\tau}$.

\textbf{NSF for Inference.}
During inference, the pre-trained NSF operates in a query-driven manner to parse scenes and predict future states. The process consists of two sequential steps:

\textit{Scene Parsing for Implicit BEV Map Initialization.} A 2D query grid $\mathcal{G}$ is defined in the ego-centric coordinate frame. For each point $(x, y)$ in $\mathcal{G}$, the semantic query $\mathbf{q}_1 = (x, y, \Delta t=0)$ is processed by NSF to output a semantic probability distribution $\mathbf{s} \in \mathbb{R}^{6}$. Evaluating these queries across $\mathcal{G}$ constructs an implicit BEV semantic map $\mathcal{M}_{\text{BEV}}$, providing the foundational scene representation.

\textit{Entity-centric Behavior and Risk Prediction.} Dynamic entities are identified from $\mathcal{M}_{\text{BEV}}$. For each entity $i$ at position $(x_i, y_i)$ with type embedding $\boldsymbol{\tau}^{(i)}$, a behavior query $\mathbf{q}_2 = (x_i, y_i, \Delta t, \boldsymbol{\tau}^{(i)})$ is formulated. The decoder outputs a trajectory offset $\mathbf{o} \in \mathbb{R}^{2}$ and TTC distribution parameters $(\mu, \log \sigma^2)$. The RiskValue $R$ is computed as:
\begin{equation}
\label{eqRP}
R = p \cdot \frac{1}{\mu + \delta}, \quad \text{where} \quad p = \frac{1}{1 + \sigma^2},
\end{equation}
and $\delta = 0.1$ ensures numerical stability.

\begin{figure}[htbp]
  \centering
  \includegraphics[width=0.72\linewidth]{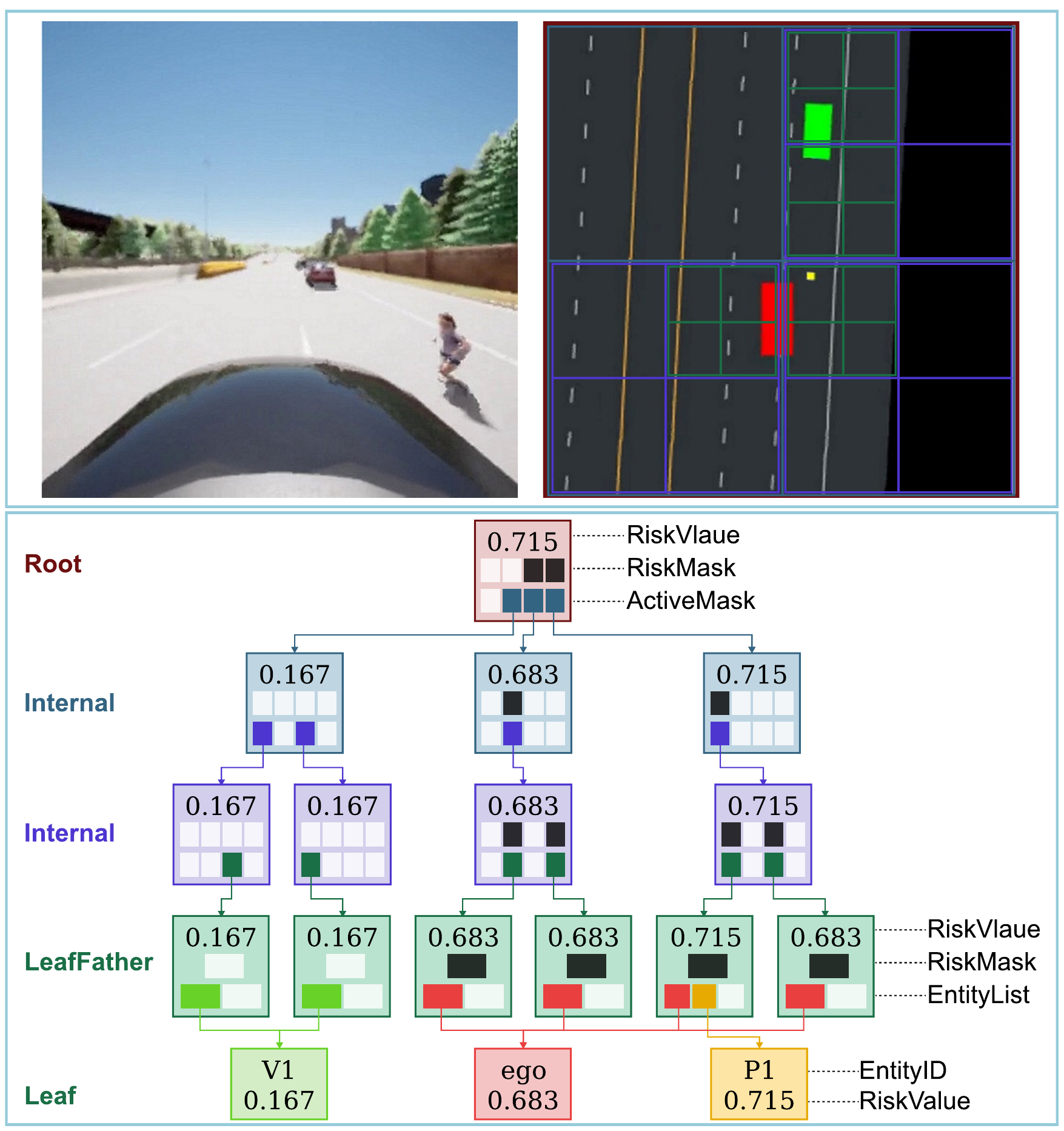}
  \caption{\textbf{The Structure of HRPT.} The quadtree organizes spatial risk information across five hierarchical layers (Root, two Internal, LeafFather, Leaf). The BEV view illustrates the spatial partitioning, while the tree diagram details the data (RiskValue, RiskMask, ActiveMask, EntityList, EntityID) stored at each node type, facilitating efficient parallel risk aggregation.} 
  \label{fig:hrpt_structure} 
\end{figure}

\begin{figure}[htbp]
  \centering
  \includegraphics[width=\linewidth]{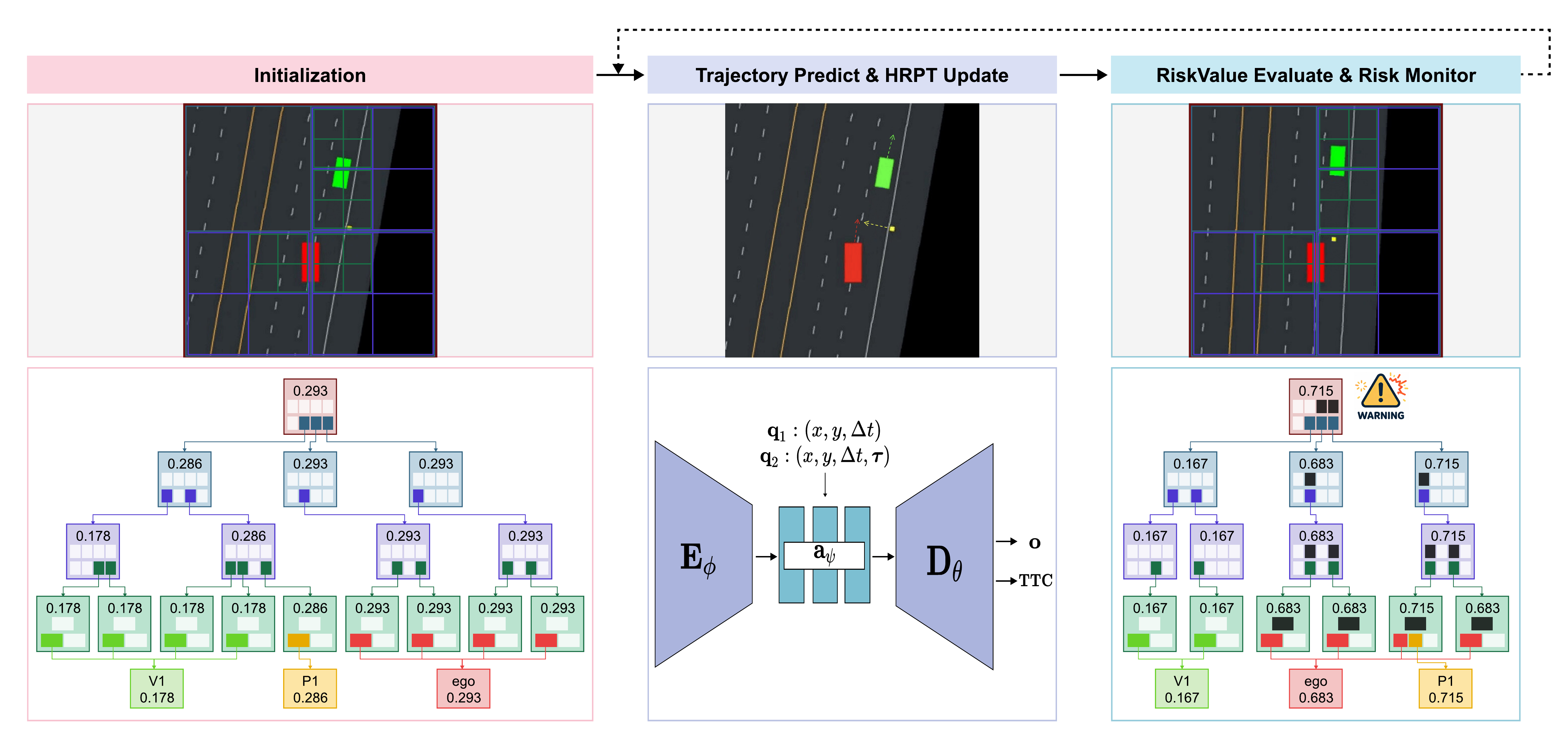}
  \caption{\textbf{The Inference Process of HRPT.} The diagram contrasts the tree's state during initial safe conditions versus upon detecting a high-risk entity. The BEV views are for conceptual illustration; the framework operates on a logical BEV representation from the NSF without requiring explicit rendering. \textbf{NOTE:} The BEV visualization is a schematic for clarity; the actual algorithm uses an implicit, query-based spatial representation.}
\label{fig:hrpt_inference}
\end{figure}

\subsection{Hierarchical Risk Perception Tree (HRPT)}

The \textbf{Hierarchical Risk Perception Tree (HRPT)} is a parallel quadtree structure that organizes dynamic entities and propagates risk information hierarchically. At inference, it leverages the pre-trained NSF to transform multi-agent risk evaluation into an efficient, parallel reasoning process for real-time safety assessment.

\subsubsection{HRPT Structure and Node Definitions}

The HRPT organizes spatial information across five hierarchical layers within the egocentric world coordinate system: (1) Root node, (2) two Internal node layers, (3) LeafFather node layer, and (4) Leaf node layer. Each node type employs specialized encoding schemes to efficiently capture and propagate risk information, with the complete structure and a detailed example of node data presented in Figure~\ref{fig:hrpt_structure}.

\textbf{Leaf Nodes} represent individual dynamic entities and store:
\begin{itemize}
    \item $\text{EntityID}$: A unique identifier for each dynamic entity, which primarily encompasses vehicles and pedestrians. Vehicle entities are further categorized into the \textit{ego vehicle} (assigned the fixed identifier \texttt{ego}) and \textit{other vehicles} (assigned identifiers starting from \texttt{V1}). Pedestrian entities are assigned identifiers starting from \texttt{P1}.
    \item $\text{RiskValue} \in \mathbb{R}^{+}$: Risk value computed from NSF queries.
    \item $\text{EntityInfo}$: State information including coordinates $(x, y)$ and type embedding $\boldsymbol{\tau}$.
\end{itemize}

\textbf{LeafFather Nodes} aggregate spatial information within fixed-size grid cells and maintain:
\begin{itemize}
    \item $\text{RiskValue}$: Maximum risk among child nodes, $R_{\text{parent}} = \max(\{ R_e \})$, where $\{ R_e \}$ represents the set of $\text{RiskValue}$s from all its children (Leaf nodes).
    \item $\text{RiskMask} \in \{0, 1\}$: Binary indicator of significant risk, computed as:
    \begin{equation}
        \text{RiskMask} = \begin{cases} 
        1 & \text{if } R_{\text{parent}} > R_{\text{Th}} \wedge p_{\text{max}} > p_{\text{Th}} \\
        0 & \text{otherwise},
        \end{cases}
    \end{equation}
     where $R_{\text{parent}}$ is the aggregated risk value, $p_{\text{max}}$ is the maximum TTC confidence probability $p$ (as defined in Eq. (\ref{eqRP})) among child nodes, $R_{\text{Th}}=0.3$ is the RiskValue threshold, and $p_{\text{Th}}=0.7$ is the confidence probability threshold.
    \item $\text{EntityList}$: Catalog of contained entities
\end{itemize}
The RiskValue of entities(leaf nodes) under each LeafFather node can be computed in parallel.

\textbf{Internal \& Root Nodes} facilitate hierarchical risk propagation through:
\begin{itemize}
    \item $\text{RiskValue}$: Maximum risk among children
    \item $\text{RiskMask} \in \{0,1\}^4$: 4-bit risk indicator per child quadrant
    \item $\text{ActiveMask} \in \{0,1\}^4$: 4-bit entity presence indicator
\end{itemize}
All four child quadrants at each Internal and Root level can be processed concurrently for efficient risk aggregation.

\subsubsection{Initialization}

The initialization process establishes the foundational state for iterative prediction. The procedure commences with BEV semantic map construction through NSF-based scene parsing of historical image sequences $\mathcal{X}$. Dynamic entities are detected from $\mathcal{M}_{\text{BEV}}$, and corresponding Leaf nodes are instantiated with initial state information.

Each entity's initial risk assessment is obtained through NSF behavior queries $\mathbf{q}_2 = (x, y, \Delta t=0, \boldsymbol{\tau})$, yielding TTC parameters $(\mu, \log \sigma^2)$ from which the initial risk value $R$ and confidence probability $p$ are derived. The spatial registration phase assigns Leaf nodes to appropriate LeafFather nodes based on coordinate information, establishing initial parent-child relationships.

The initialization culminates with bottom-up aggregation where risk values and mask encodings propagate upward through the hierarchy. LeafFather nodes compute aggregated RiskValues and binary RiskMasks, while Internal and Root nodes determine their states through maximum risk aggregation and bitwise mask operations.

\subsubsection{Iterative Prediction Protocol}

The HRPT executes $T=50$ iterative prediction steps ($\delta t = 0.1s$), each comprising five distinct phases, with the complete workflow illustrated in Figure~\ref{fig:hrpt_inference}.

\textbf{Trajectory Predict:} Entity trajectories are updated through NSF queries:
\begin{equation}
    \mathbf{p}_{\text{new}} = \mathbf{p}_{\text{current}} + \mathbf{o},
\end{equation}
where $\mathbf{p}_{\text{current}}$ denotes the entity's current position and $\mathbf{p}_{\text{new}}$ represents its predicted future position.

\textbf{HRPT Update:} Leaf node coordinates are adjusted with spatial reassignment to maintain accurate registration.

\textbf{RiskValue Evaluate:} Updated TTC parameters $(\mu, \log \sigma^2)$ are obtained through NSF behavior queries at new positions. Bottom-up risk aggregation recalculates all nodal RiskValues and masks according to equation \eqref{eqRP}, where $R$ represents the RiskValue, $p$ denotes the confidence probability of the predicted Time-to-Collision.

\textbf{Risk Monitor:} Root node's RiskValue and RiskMask are monitored for threshold violations, with risk source identification through hierarchical mask traversal.

This structured iterative process enables predictive, hierarchical risk assessment that anticipates potential collision scenarios and supports verifiable navigation decisions.

\subsection{Sim2Real Enhancement}
\label{sec:sim2real}

While trained exclusively on synthetic CARLA data, the NSF framework requires robust evaluation on real-world datasets. To address the visual domain gap at inference without retraining, we integrate geometric and semantic priors from pre-trained foundation models into the frozen NSF architecture.

\textbf{Input Enhancement via Early-Fusion.}
For a real-world image sequence $\mathcal{X} \in \mathbb{R}^{T \times H \times W \times 3}$, we obtain a depth map $\mathcal{D} \in \mathbb{R}^{T \times H \times W \times 1}$ and semantic features $\mathcal{S} \in \mathbb{R}^{T \times H \times W \times C_s}$ from pre-trained models. These are concatenated and projected via a fixed linear transformation $\mathcal{P}$ to maintain original input dimensions:
\begin{equation}
\mathcal{X}_{\text{enhanced}} = \mathcal{P}([\mathcal{X}, \mathcal{D}, \mathcal{S}]) \in \mathbb{R}^{T \times H \times W \times 3}.
\end{equation}
This enhanced input is fed to the frozen encoder $\mathbf{E}_{\phi}$.

\textbf{Feature Enhancement via Mid-Fusion.}
The original image sequence is processed by $\mathbf{E}_{\phi}$ to extract initial features $\mathbf{F}_{\text{feat}} \in \mathbb{R}^{T \times P \times C}$. External models provide depth and semantic features $\mathbf{F}_d, \mathbf{F}_s \in \mathbb{R}^{T \times P \times C}$, which are aligned spatially and temporally. These are fused through a weighted summation:
\begin{equation}
\mathbf{F}_{\text{enhanced}} = \mathbf{F}_{\text{feat}} + \alpha_d \mathbf{F}_d + \alpha_s \mathbf{F}_s,
\end{equation}
where $\alpha_d=0.5, \alpha_s=0.3$ are fixed scaling coefficients. The enhanced features $\mathbf{F}_{\text{enhanced}} \in \mathbb{R}^{T \times P \times C}$ proceed through the remaining NSF components.

Both methods require no additional training and enhance the NSF's robustness to real-world visual variations while preserving its pre-trained parameters and architectural integrity.

\section{EXPERIMENT}

\subsection{Experimental Setup}

\subsubsection{Simulation environment}
\label{sec:carla}

We employ the CARLA simulator \cite{pmlr-v78-dosovitskiy17a} with the SafeBench framework \cite{NEURIPS2022_a48ad12d} for closed-loop autonomous vehicle evaluation. Our evaluation encompasses 8 critical traffic scenarios from the NHTSA Pre-Crash Typology Report \cite{najm2007precrash} (including Straight Obstacle, Lane Changing, etc.), selected following the methodology in \cite{zhang2024chatscene, liu2025adversarial}, totaling 800 challenging scenarios (10 routes × 10 variations per scenario) for comprehensive benchmarking \cite{liu2025adversarial}.

The BEV representation is configured with 512×512 pixel images covering a 32m×32m physical area, providing 12m rearward and 20m forward field of view from the ego vehicle through a 64-pixel viewpoint offset. We generate precise ground truth data including semantic segmentation masks, entity trajectories, and Time-to-Collision (TTC) values representing the duration to actual collision. This comprehensive dataset enables reproducible training and evaluation of safety-critical scenario detection in our risk assessment framework.

\subsubsection{Real-World Datasets}
Our model is validated on two standard benchmarks for accident anticipation. 

\textbf{DAD.} The Dashcam Accident Dataset (DAD)~\cite{chan2016anticipating} consists of 1,750 five-second, 20-FPS video clips of urban traffic, where accidents in its 620 positive samples are confined to the final 10 frames. 

\textbf{CCD.} The larger Car Crash Dataset (CCD)~\cite{bao2020uncertainty} offers greater environmental diversity across its 4,500 videos, supported by rich metadata. Its 1,500 accident instances occur within the last two seconds and are augmented by 3,000 normal driving clips from BDD100K~\cite{bdd100k}.

\subsubsection{Evaluation Metrics}
\label{sec:metrics}
We evaluate our model using three key metrics: \textbf{Average Precision (AP)} , \textbf{Mean Time-to-Accident (mTTA)} , and \textbf{Accident Object Localization Accuracy (AOLA)} \cite{liao2024and} for spatial precision.

\textbf{AP} measures accident detection accuracy under natural class imbalance, calculated as the area under the Precision-Recall curve.

\textbf{mTTA} measures the average time between the first valid warning and accident occurrence, we employ a rigorous criterion to determine the first valid warning moment $t_\theta$. At each timestamp $t_\theta$, our model performs efficient iterative reasoning to predict the RiskValue for future time $t_\theta + \delta t$. While the first moment the predicted RiskValue exceeds a threshold could be used, we define $t_\theta$ as a valid warning moment only when: (1) the RiskValue predicted for $t_\theta + \delta t$ first crosses the threshold, and (2) the absolute error between the predicted TTC ($\hat{T}_{t_\theta + \delta t}$) and ground-truth TTC ($T_{GT, t_\theta + \delta t}$) satisfies $|\hat{T}_{t_\theta + \delta t} - T_{GT, t_\theta + \delta t}| \leq \epsilon$ (with $\epsilon = 0.3$s based on autonomous driving requirements). The mTTA is then computed as $\frac{1}{N}\sum_{i=1}^{N} (t_{\text{collision}, i} - t_{\theta, i})$ across all positive samples.

\textbf{AOLA} evaluates the model's precision in identifying accident-related objects, computed as the fraction of correctly predicted objects among all detected objects across frames.

\subsubsection{Baselines}
We compare our model against key state-of-the-art methods representing major technical trends. This includes attention-based models (DSA~\cite{chan2016anticipating}, adaLEA~\cite{suzuki2018anticipating}, DSTA~\cite{karim2022dynamic}), graph-based frameworks (Ustring~\cite{bao2020uncertainty}, GG~\cite{thakur2024graph}), and the recent LLM-based approach of Liao et al.~\cite{liao2024and} that predicts "when, where, and what". 

\subsubsection{Implementations}

All experiments are conducted on a server equipped with an Intel(R) Core(TM) i9-14900K CPU and two NVIDIA GeForce RTX 4090 GPUs, each with 24GB of memory.

Our transformer architecture employs $L=2$ layers with 4 parallel attention heads. For encoder selection, we evaluate ResNet-18, ResNet-34, and ResNet-50, with ResNet-34 delivering the optimal performance.

To address the absence of BEV trajectory and semantic annotations in real-world datasets (DAD and CCD), we utilize the simulated scenarios from Section~\ref{sec:carla}, processed into video frames for model training. Each scenario is partitioned into training and testing subsets with a 4:1 ratio. This framework facilitates cross-domain evaluation, where models are trained on synthetic data and evaluated on real-world benchmarks. We maintain the official training/testing partitions for DAD (1,284/466) and CCD (3,600/900), using these datasets exclusively for testing purposes.

\subsection{Comparison to State-of-the-art (SOTA)}

We conduct comprehensive evaluations to benchmark our proposed framework against SOTA methods across both simulated and real-world datasets. The comparative analysis validates the effectiveness of our approach in accurate risk anticipation and demonstrates its potential for sim-to-real transfer.

\begin{table}[htbp]
\centering
\caption{Performance Comparison on the CARLA Simulation Benchmark.}
\label{tab:carla}

\setlength{\tabcolsep}{3pt}

\resizebox{\textwidth}{!}{
\begin{tabular}{@{}c|c|cccccccc|c@{}}
\Xhline{1pt}
\noalign{\vskip 1.5pt}
\toprule
\multirow{2}{*}[-1ex]{\textbf{Metric}} & \multirow{2}{*}[-1ex]{\textbf{Algo.}} & \multicolumn{8}{c|}{\textbf{Base Traffic Scenarios}} & \multirow{2}{*}[-1ex]{\textbf{Avg.}} \\
 &
   &
  \begin{tabular}[c]{@{}c@{}}\scriptsize Straight\\ \scriptsize Obstacle\end{tabular} &
  \begin{tabular}[c]{@{}c@{}}\scriptsize Turning\\ \scriptsize Obstacle\end{tabular} &
  \begin{tabular}[c]{@{}c@{}}\scriptsize Lane\\ \scriptsize Changing\end{tabular} &
  \begin{tabular}[c]{@{}c@{}}\scriptsize Vehicle\\ \scriptsize Passing\end{tabular} &
  \begin{tabular}[c]{@{}c@{}}\scriptsize Red-light\\ \scriptsize Running\end{tabular} &
  \begin{tabular}[c]{@{}c@{}}\scriptsize Unprotected\\ \scriptsize Left-turn\end{tabular} &
  \begin{tabular}[c]{@{}c@{}}\scriptsize Right-\\ \scriptsize turn\end{tabular} &
  \begin{tabular}[c]{@{}c@{}}\scriptsize Crossing\\ \scriptsize Negotiation\end{tabular} &
   \\ \midrule
\multirow{3}{*}{AP(\%)}              & DSA                             & 65    & 75    & 80    & 75    & 70    & 70    & 80    & 75   & 73.8                              \\
                                 & Liao et al.                     & 80    & \textbf{85}    & \textbf{85}    & \textbf{80}    & 80    & 75    & 80    & 70   & 79.3                              \\ 
                                 & \textbf{Ours}                   & \textbf{90}    & \textbf{85}    & \textbf{85}    & \textbf{80}    & \textbf{90}    & \textbf{80}    & \textbf{85}    & \textbf{80}   & \textbf{84.3}                              \\ \midrule
\multirow{3}{*}{mTTA}            & DSA                             & 2.58    & 2.26    & 2.52    & 2.27    & 2.65    & 2.13    & 2.57    & 2.37   & 2.42                              \\
                                 & Liao et al.                     & 3.22    & 3.11    & 3.29    & 2.93    & 3.24    & 3.16    & 3.49    & 3.27   & 3.21                              \\ 
                                 & \textbf{Ours}                   & \textbf{3.68}    & \textbf{3.81}    & \textbf{3.49}    & \textbf{3.3}    & \textbf{3.77}    & \textbf{3.21}    & \textbf{3.54}    & \textbf{3.52}   & \textbf{3.54}                              \\ \midrule
\multirow{2}{*}{AOLA}            & Liao et al.                     & 0.86    & 0.84    & 0.85    & 0.84    & 0.87    & 0.83    & 0.84    & 0.81   & 0.84                              \\ 
                                 & \textbf{Ours}                   & \textbf{0.91}    & \textbf{0.89}    & \textbf{0.92}    & \textbf{0.89}    & \textbf{0.92}    & \textbf{0.88}    & \textbf{0.92}    & \textbf{0.9}   & \textbf{0.9}                              \\ 

\bottomrule
\noalign{\vskip 1.5pt}
\Xhline{1.5pt}
\end{tabular}
}
\end{table}

\begin{table}[htbp]
    \centering
    \caption{Performance Comparison on the DAD and CCD Real-Wrold Datasets.}
    \label{tab:real}
    \begin{tabular}{@{}l ccc cc@{}}
        \Xhline{1.5pt}
        \noalign{\vskip 1.5pt}
        \toprule
        \multirow{2}{*}{Model} & \multicolumn{3}{c}{DAD} & \multicolumn{2}{c}{CCD} \\
        \cmidrule(lr){2-4} \cmidrule(lr){5-6}
        & AP(\%)$\uparrow$ & mTTA(s)$\uparrow$ & AOLA$\uparrow$ & AP(\%)$\uparrow$ & mTTA(s)$\uparrow$ \\
        \midrule
        DSA~\cite{chan2016anticipating}         & 48.1 & 1.34 & - & 98.7 & 3.08 \\
        adaLEA~\cite{suzuki2018anticipating}    & 52.3 & 3.43 & - & 99.2 & 3.45 \\
        Ustring~\cite{bao2020uncertainty}       & 53.7 & 3.53 & - & 99.5 & 3.74 \\
        DSTA~\cite{karim2022dynamic}            & 56.1 & 3.66 & - & 99.6 & 3.87 \\
        Liao et al.~\cite{liao2024and}          & \textbf{69.2} & \underline{4.26} & \textbf{0.89} & 99.7 & 3.93 \\
        GG~\cite{thakur2024graph}               & 63.6 & \textbf{4.45} & - & \textbf{99.9} & \textbf{4.96} \\
        \midrule
        \textbf{Ours}                           & 56.7 & 3.72 & 0.65 & 99.2 & 3.67\\
        \textbf{Ours + EF}                      & 67.2 & 4.13 & \underline{0.85} & \underline{99.8} & 3.87\\
        \textbf{Ours + MF}                      & \underline{68.5} & 4.23 & \textbf{0.89} & \textbf{99.9} & \underline{3.95}\\
        \bottomrule
        \noalign{\vskip 1.5pt}
        \Xhline{1.5pt}
    \end{tabular}
    \vspace{0.3em}
    
    \footnotesize
    The best and second-best results per column are in \textbf{bold} and \underline{underlined}, respectively; EF/MF denotes Early/Mid-Fusion, while “-” indicates unavailable values.
\end{table}

\textbf{Evaluation on CARLA Simulation.}
We first evaluate our method on the CARLA simulation scenarios described in Section~\ref{sec:carla}. To ensure a fair comparison with baseline methods \textbf{DSA}~\cite{chan2016anticipating} and \textbf{Liao et al.}~\cite{liao2024and}, which require first-person view object annotations, we generate corresponding ground truth labels from the simulated scenarios and adhere to identical training/testing splits. As summarized in Table \ref{tab:carla}, our method achieves superior performance across all three metrics (AP, mTTA, AOLA) in all eight base traffic scenarios, establishing a new state-of-the-art in the simulated environment. The consistent improvements demonstrate the strong scene understanding and predictive capability of our NSF-HRPT framework.

\textbf{Evaluation on Real-World Datasets.}
To assess the sim-to-real generalization ability of our model, we directly evaluate our NSF-HRPT framework, trained exclusively on synthetic CARLA data, on the real-world DAD and CCD datasets without any fine-tuning. The results in Table \ref{tab:real} show that our base model already surpasses several early baseline methods (e.g., DSA, adaLEA), indicating a promising sim-to-real transfer capability. However, a performance gap remains compared to the current SOTA methods, which we attribute to the domain shift in visual appearance and increased scene complexity (e.g., vehicle types, background variations) in real-world data.

To bridge this domain gap, we employ the Sim2Real enhancement strategies described in Section~\ref{sec:sim2real}, utilizing frozen pre-trained models for semantic segmentation~\cite{SAM23} and depth estimation~\cite{VITDP}. Incorporating Early-Fusion (EF) and Mid-Fusion (MF) brings substantial performance gains across all metrics. Notably, with Mid-Fusion enhancement, our method achieves competitive results, matching or approaching the performance of the current SOTA on several key metrics. 

\subsection{Ablation Studies}

\begin{figure}[htbp]
  \centering
  \includegraphics[width=\linewidth]{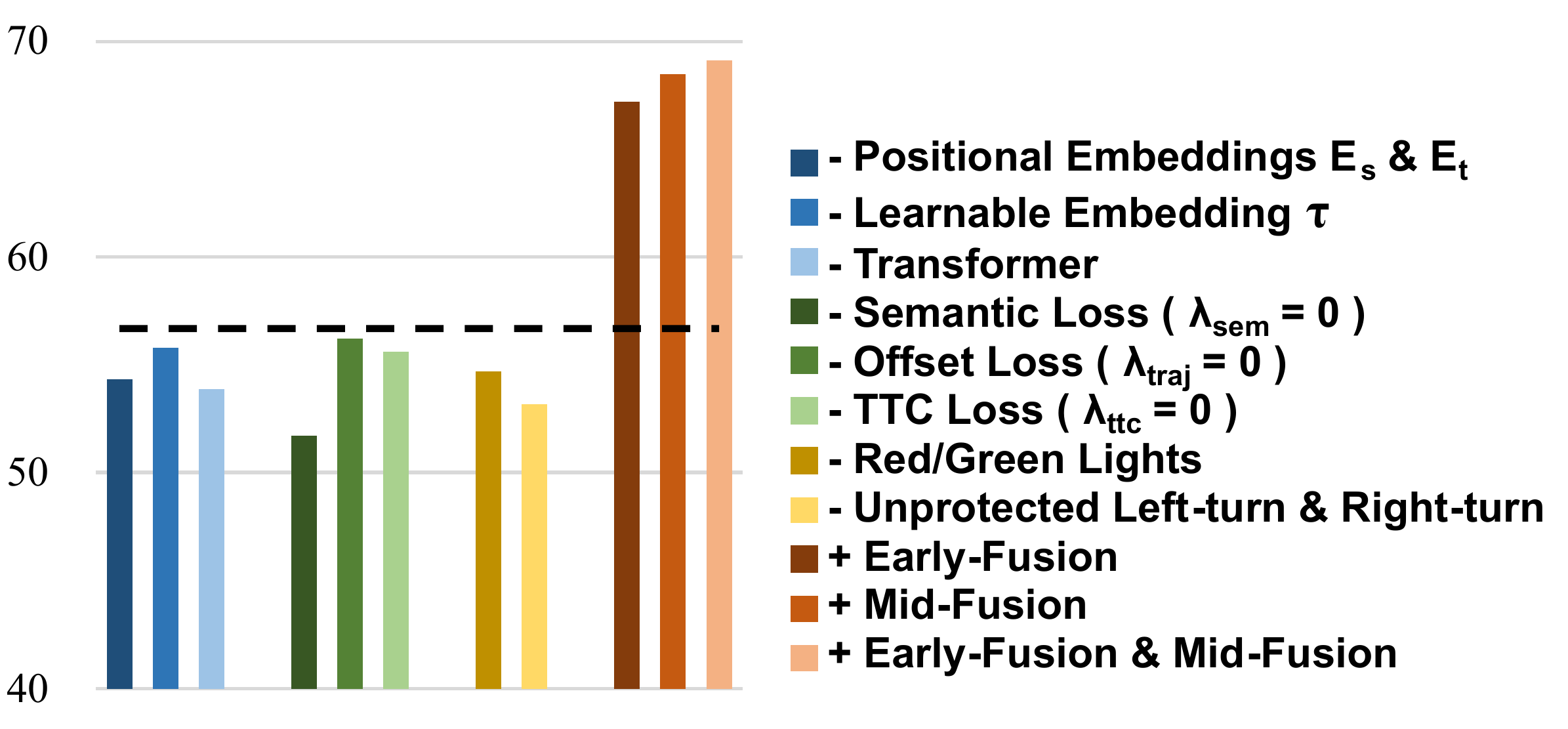}
  \caption{Component Analysis on DAD Dataset Using AP. "- Red/Green Lights": removing the 'red light' and 'green light' classes from the semantic prediction task. "- Unprotected Left-turn $\&$ Right-turn": excluding unprotected left-turn and right-turn scenarios from the training set.}
  \label{fig:AP}
\end{figure}

\begin{figure}[htbp]
  \centering
  \includegraphics[width=\linewidth]{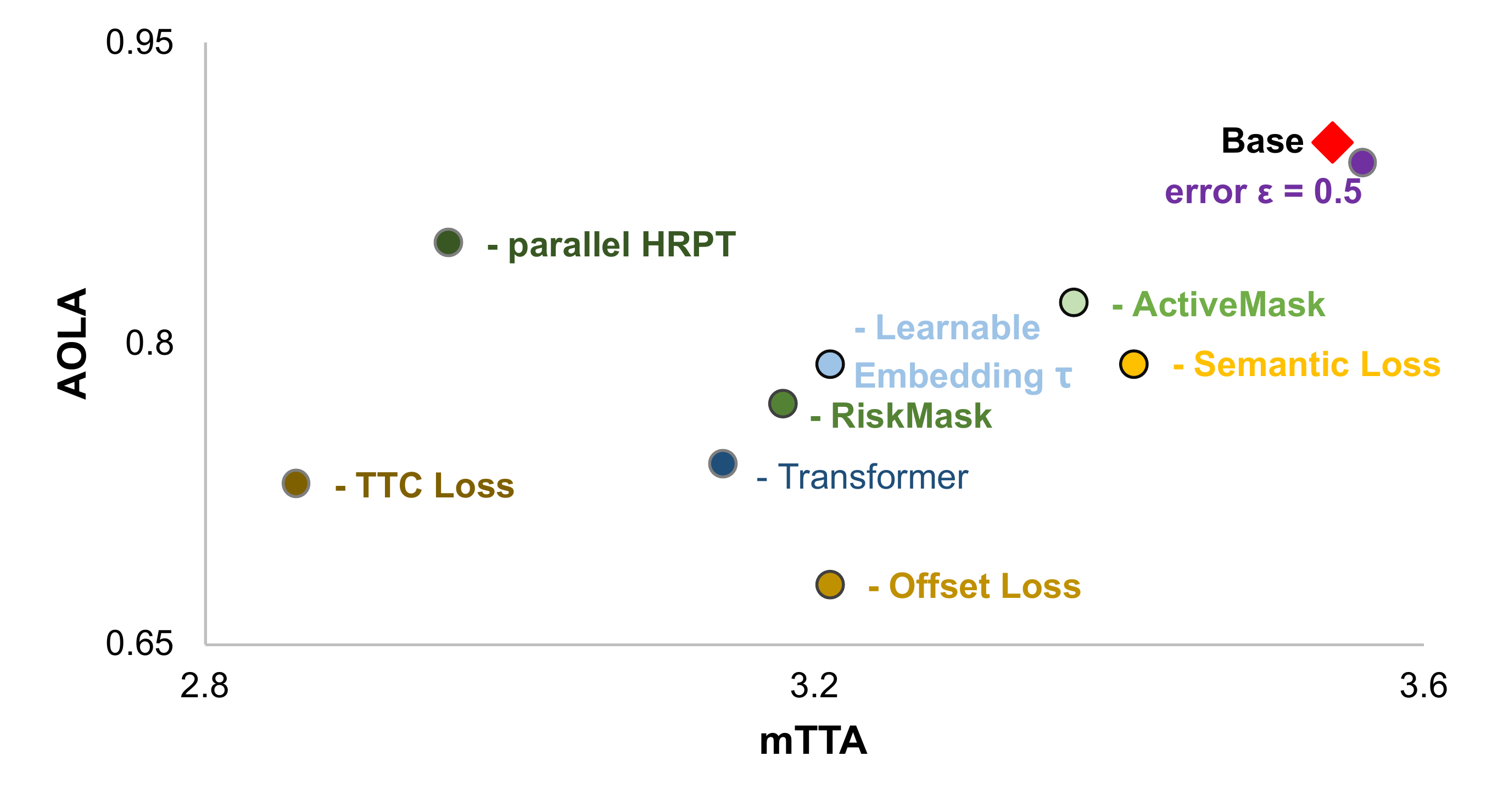}
  \caption{Component Analysis on CARLA Using mTTA and AOLA.} 
  \label{fig:carla} 
\end{figure}

We conduct systematic ablation studies to evaluate the contributions of individual components in our framework across two evaluation settings. Results are summarized in Figure~\ref{fig:AP} (DAD dataset, AP metric) and Figure~\ref{fig:carla} (CARLA simulation, mTTA and AOLA metrics).

On the DAD benchmark, we analyze core architectural components including the Positional Embeddings, Learnable Type Embedding $\boldsymbol{\tau}$, and Transformer module. The multi-task learning objectives (Semantic, Offset, and TTC Losses) and training data configuration are examined. Our Sim2Real strategy, employing frozen pre-trained models, demonstrates effective domain adaptation and enhances robustness to real-world variations.

Complementary evaluation on CARLA simulation assesses the hierarchical reasoning mechanism of the HRPT, including its parallel structure, RiskMask, and ActiveMask components. The importance of the Learnable Embedding $\boldsymbol{\tau}$, Transformer module, and multi-task losses is validated. All components contribute significantly to the final performance in both experimental settings.

\section{CONCLUSION AND FUTURE WORK}

In this work, we introduce \textbf{NSF-HRPT}, a novel framework for quantitative risk assessment, built on a dual-stage paradigm unifying learning and reasoning. At its core, an \textbf{NSF} learns continuous spatiotemporal representations from simulation data, encoding scene semantics, trajectory predictions, and probabilistic Time-to-Collision (TTC) distributions. The proposed \textbf{HRPT} enables efficient parallel risk reasoning via spatial-hierarchical computation, while our \textbf{Sim2Real} strategy boosts real-world generalizability with no retraining required. Extensive experiments demonstrate state-of-the-art performance on synthetic benchmarks, and competitive accuracy for both TTC estimation and risk localization on real-world datasets.

Future research will pursue three key directions to extend this work. First, we will develop hybrid training paradigms that integrate limited real-world data with simulation data to enhance the NSF's domain adaptation capabilities, potentially through semi-supervised or adversarial training techniques. Second, we plan to incorporate large language models (LLMs) to provide higher-level risk interpretation and reasoning, enabling more intuitive and actionable warning generation for autonomous systems. Finally, we aim to construct comprehensive real-world datasets incorporating bird's-eye-view trajectory annotations and detailed risk assessments, which will serve as essential benchmarks for advancing safety-critical perception research. These efforts will collectively strengthen the framework's robustness, interpretability, and applicability, ultimately facilitating safer and more reliable deployment in real-world autonomous driving systems.

\section*{ACKNOWLEDGMENT}
We acknowledge the use of large language models (LLMs) during the preparation of this manuscript. LLMs were used exclusively in an auxiliary capacity for Chinese-to-English translation and text polishing, LaTeX code generation, and assisting with proofreading to ensure logical consistency and formatting standardization.

This work was partly supported by Ningbo Youth Science and Technology Innovation Leading Talent Project (2024QL044) and Ningbo Key R\&D Program (2025Z047).

\bibliographystyle{unsrtnat}
\bibliography{references}

\end{document}